\documentclass[sigconf]{acmart}
\AtBeginDocument{%
  }

\setcopyright{acmlicensed}
\usepackage{enumitem}
\usepackage{natbib}
\usepackage{algorithm}
\usepackage{algorithmic}
\usepackage{bbding}
\usepackage{multirow}
\usepackage{cleveref}
\crefname{equation}{Eq.}{Eqs.}
\Crefname{equation}{Eq.}{Eqs.}
\copyrightyear{2018}
\acmYear{2018}
\acmDOI{XXXXXXX.XXXXXXX}
\newcommand{\etal}{\textit{et al.}}
\acmISBN{978-1-4503-XXXX-X/2018/06}

\begin{document}

\title{Semi-supervised Image Dehazing via Expectation-Maximization and Bidirectional Brownian Bridge Diffusion Models}

\author{Bing Liu}
\affiliation{%
  \institution{China University of Mining and Technology}
  \institution{Mine Digitization Engineering Research Center of the Ministry of Education}
  \city{XuZhou, JiangSu}
  \country{China}
}
\email{liubing@cumt.edu.cn}

\author{Le Wang}
\authornote{Both authors contributed equally to this research.}
\affiliation{%
  \institution{China University of Mining and Technology}
  \institution{Mine Digitization Engineering Research Center of the Ministry of Education}
  \city{XuZhou, JiangSu}
  \country{China}
}
\email{TS23170132P31@cumt.edu.cn}

\author{Mingming Liu}
\affiliation{%
  \institution{China University of Mining and Technology}
  \city{XuZhou, JiangSu}
  \country{China}
}

\author{Hao Liu}
\affiliation{%
  \institution{China University of Mining and Technology}
  \institution{Mine Digitization Engineering Research Center of the Ministry of Education}
  \city{XuZhou, JiangSu}
  \country{China}
}

\author{Rui Yao}
\affiliation{%
  \institution{China University of Mining and Technology}
  \institution{Mine Digitization Engineering Research Center of the Ministry of Education}
  \city{XuZhou, JiangSu}
  \country{China}
}

\author{Yong Zhou}
\affiliation{%
  \institution{China University of Mining and Technology}
  \institution{Mine Digitization Engineering Research Center of the Ministry of Education}
  \city{XuZhou, JiangSu}
  \country{China}
}

\author{Peng Liu}
\affiliation{%
 \institution{China University of Mining and Technology}
  \institution{Mine Digitization Engineering Research Center of the Ministry of Education}
  \city{XuZhou, JiangSu}
  \country{China}
}

\author{Tongqiang Xia}
\affiliation{%
  \institution{China University of Mining and Technology}
  \institution{Mine Digitization Engineering Research Center of the Ministry of Education}
  \city{XuZhou, JiangSu}
  \country{China}}

\renewcommand{\shortauthors}{Le Wang et al.}

\begin{abstract}
Existing dehazing methods deal with real-world haze images with difficulty, especially scenes with thick haze. One of the main reasons is lacking real-world pair data and robust priors. To avoid the costly collection of paired hazy and clear images, we propose an efficient Semi-supervised Image Dehazing via Expectation-Maximization and Bidirectional Brownian Bridge Diffusion Models (EM-B\textsuperscript{3}DM) with a two-stage learning scheme. In the first stage, we employ the EM algorithm to decouple the joint distribution of paired hazy and clear images into two conditional distributions, which are then modeled using a unified Brownian Bridge diffusion model to directly capture the structural and content-related correlations between hazy and clear images. In the second stage, we leverage the pre-trained model and large-scale unpaired hazy and clear images to further improve the performance of image dehazing. Additionally, we introduce a detail-enhanced Residual Difference Convolution block (RDC) to capture gradient-level information, significantly enhancing the model’s representation capability. Extensive experiments demonstrate that our EM-B\textsuperscript{3}DM achieves superior or at least comparable performance to state-of-the-art methods on both synthetic and real-world datasets.
\end{abstract}






\maketitle

\section{Introduction}
Haze is caused by the scattering effect of aerosol particles in the atmosphere, leading to photographs captured in hazy weather typically exhibiting low contrast and fidelity \cite{c:1}. Dehazing methods aim to remove haze and enhance the contrast and color integrity of real-world hazy images, which plays a crucial role in enabling computer vision tasks such as image segmentation \cite{c:2, c:3} and object detection \cite{c:4} under hazy weather conditions. The degradation caused by haze can be described by the Atmospheric Scattering Model (ASM) \cite{c:5}:
\begin{equation}
  I(p) = J(p) e^{-\beta d(p)} + A (1 - e^{-\beta d(p)}),
  \label{eq:1}
\end{equation}
where $J(p)$ and $I(p)$ indicate the p-th pixel position of hazy and clear image, respectively, and $A$ is the global atmosphere light. The transmission map $e^{-\beta d(p)}$ is defined by the scene depth $d(p)$ and the scattering coefficient $\beta$, which reflects the haze density. 

Existing supervised learning methods for image dehazing based on deep learning, which either rely on physical models \cite{c:6,c:7,c:8} or directly learn clear images \cite{c:9,c:10,c:11,c:12,c:13}, have demonstrated impressive results on specific test sets by training on extensive synthetic hazy-clear image pairs. However, a pronounced domain gap exists between synthetic and real-world hazy images. Collecting a large number of ideal hazy and clear image pairs is not only a time-consuming and resource-intensive task but may also present an almost insurmountable challenge. 

Aiming at extracting dehazing cues from unpaired training data, numerous un-/semi-supervised deep learning methods have emerged in recent years. Among unsupervised learning approaches, the implementation of dehazing and rehazing cycles has attracted considerable attention \cite{c:14,c:15,c:19}, as it offers a straightforward and effective strategy for maintaining content consistency throughout the process of domain transformation. Meanwhile, semi-supervised methods enhance the dehazing performance by combining synthetic and real data. However, it is insufficient to directly inherit the CycleGAN framework from unpaired image-to-image translation methods for utilizing unpaired hazy and clear images, since the training process of GANs is often unstable and suffers from mode collapse. In contrast, diffusion models \cite{c:21,c:22,c:23} have shown superior performance in modeling data distributions compared to GAN-based models \cite{c:24}. The Brownian Bridge Diffusion Model (BBDM) \cite{c:59} utilizes the stochastic properties of the Brownian Bridge process, achieving strong performance in image translation. However, limited by the adopted KL divergence of marginal distributions $KL(q(x)||p(x))$, BBDM fails to implement bidirectional translation with a single model to maintain structural consistency and requires a large amount of paired data for training.

In this paper, we propose a Semi-supervised Image Dehazing via Expectation-Maximization and Bidirectional Brownian Bridge Diffusion Models, termed as EM-B\textsuperscript{3}DM. Compared to existing diffusion methods, we provide a clear theoretical guarantee that the final diffusion result can generate the desired conditional distribution without relying on a large amount of paired data for training. The proposed framework is optimized by a paired training stage and an unpaired training stage. In the first stage, we utilize the EM algorithm to decouple the joint distribution of paired hazy and clear images into two conditional distributions, aiming to perform dehazing or haze generation via a conditional generative model. We employ a unified Brownian Bridge diffusion model to effectively learn these two conditional distributions. In the second stage, we leverage the pre-trained model in the first stage to perform unsupervised image dehazing with a large number of unpaired hazy and clear images, further improving the generalization performance of image dehazing. Additionally,we develop a detail-enhanced Residual Difference Convolution block (RDC) to capture gradient-level information, which can effectively improve the model's representation capability.

Overall, our contributions can be summarized as follows:
\begin{itemize}[topsep=0pt,partopsep=0pt,itemsep=0pt]
\item We present a novel semi-supervised image dehazing framework that seamlessly integrates the EM algorithm and bidirectional Brownian Bridge
diffusion models to perform image dehazing with limited paired data.
\item We develop a detail-enhanced Residual Difference Convolution block  that effectively captures gradient-level information to further improve the representation and generalization capabilities of EM-B\textsuperscript{3}DM.
\item We implement and extensively evaluate the proposed EM-B\textsuperscript{3}DM on widely-used benchmark datasets. The qualitative and quantitative experiments demonstrate its superior performance compared to the state-of-the-art methods. 
\end{itemize}

\section{Related Works}
This section briefly reviews previous works closely related to ours, grouped into supervised and un-/semi-supervised  methods, diffusion models, and Brownian Bridge.

\subsection{Supervised learning approaches}
The supervised learning methods leverage the advancements in deep learning for image dehazing. Early approaches generated haze-free images by estimating the transmission map and global atmospheric light in the atmospheric scattering model. For instance, MSCNN \cite{c:32} estimates the transmission map and restores clean results, while AOD-Net \cite{c:33} directly produces haze-free images in an end-to-end manner. DehazeFormer \cite{c:34} introduces Vision Transformers \cite{c:36} to replace convolutional neural network-based methods. TC-Net \cite{c:35} employs a compact dehazing network based on an autoencoder-like architecture, enhancing the network's feature representation capability through a feature enhancement module and preserving detail information via an attention fusion module. Although supervised methods achieve ideal performance on synthetic datasets, they are prone to overfitting to the training data, leading to poor generalization ability.

\subsection{Un-/Semi-supervised learning approaches}
Un-/Semi-supervised image dehazing methods gain attention recently due to their independence from paired hazy and clear images for training, addressing the challenge of acquiring matched data in real-world settings. Li \etal \cite{c:52} propose an effective semi-supervised learning algorithm for single image dehazing that combines supervised and unsupervised branches, enabling training on synthetic data while generalizing well to real-world scenarios. The semi-supervised dehazing algorithm SADnet \cite{c:53} trains on both synthetic datasets and real hazy images, enhancing its representational power through a Channel-Spatial Self-Attention mechanism. Zhang \etal \cite{c:54} formulate the dehazing task as a semi-supervised domain translation problem and design two auxiliary domain translation tasks to reduce the domain gap between synthetic and real hazy images. Dong \etal \cite{c:55} introduce a semi-supervised learning approach that combines synthetic data and real images for single image dehazing training by incorporating a domain alignment module and a haze-aware attention module. 

CycleGAN-based frameworks are key in unpaired dehazing. Cycle-Dehaze \cite{c:15} enhances CycleGAN by integrating perceptual loss and cycle-consistency, eliminating the need for atmospheric scattering model parameters. CDNet \cite{c:27} uses an encoder-decoder architecture to estimate object-level transmission maps for haze-free scene recovery. DD-CycleGAN  introduces a novel double-discriminator framework within a cycle-consistent generative adversarial network. D4 \cite{c:29} presents a self-augmented image dehazing paradigm, which enhances the model's generalization capability by leveraging the scattering coefficient and depth information inherent in both hazy and clear images.
\subsection{Brownian Bridge} Brownian Bridge is a stochastic process that represents a continuous path constrained to start and end at specific points. It is derived from Brownian motion, characterized by continuous trajectories and stationary increments. In this process, increments follow a normal distribution with a mean of zero, making it useful for statistical applications such as hypothesis testing and confidence interval construction. The variance is formulated as: 
\begin{equation}
  [B(t)]=\frac{t(T-t)}{T},
  \label{eq:5}
\end{equation}
where $T$ is the total duration, reflecting the dispersion of values over time. Brownian Bridge is a valuable tool for modeling dynamic systems with specified boundary conditions. 

In our EM-B\textsuperscript{3}DM, the Brownian Bridge mechanism serves dual purposes: 1) It constraints the diffusion trajectory between hazy observations and latent clean images through terminal pinning, preventing divergence in the high-dimensional image space; 2) The analytical variance structure enables physics-compatible noise modeling that adapts to spatially variant haze concentrations. Unlike conventional diffusion models with monotonically increasing variance, bridge-based approach achieves better trade-off between exploration and exploitation through its parabolic uncertainty profile, particularly effective for handling non-uniform haze distributions.

\begin{figure*}[t]
\centering
\includegraphics[width=1\textwidth]{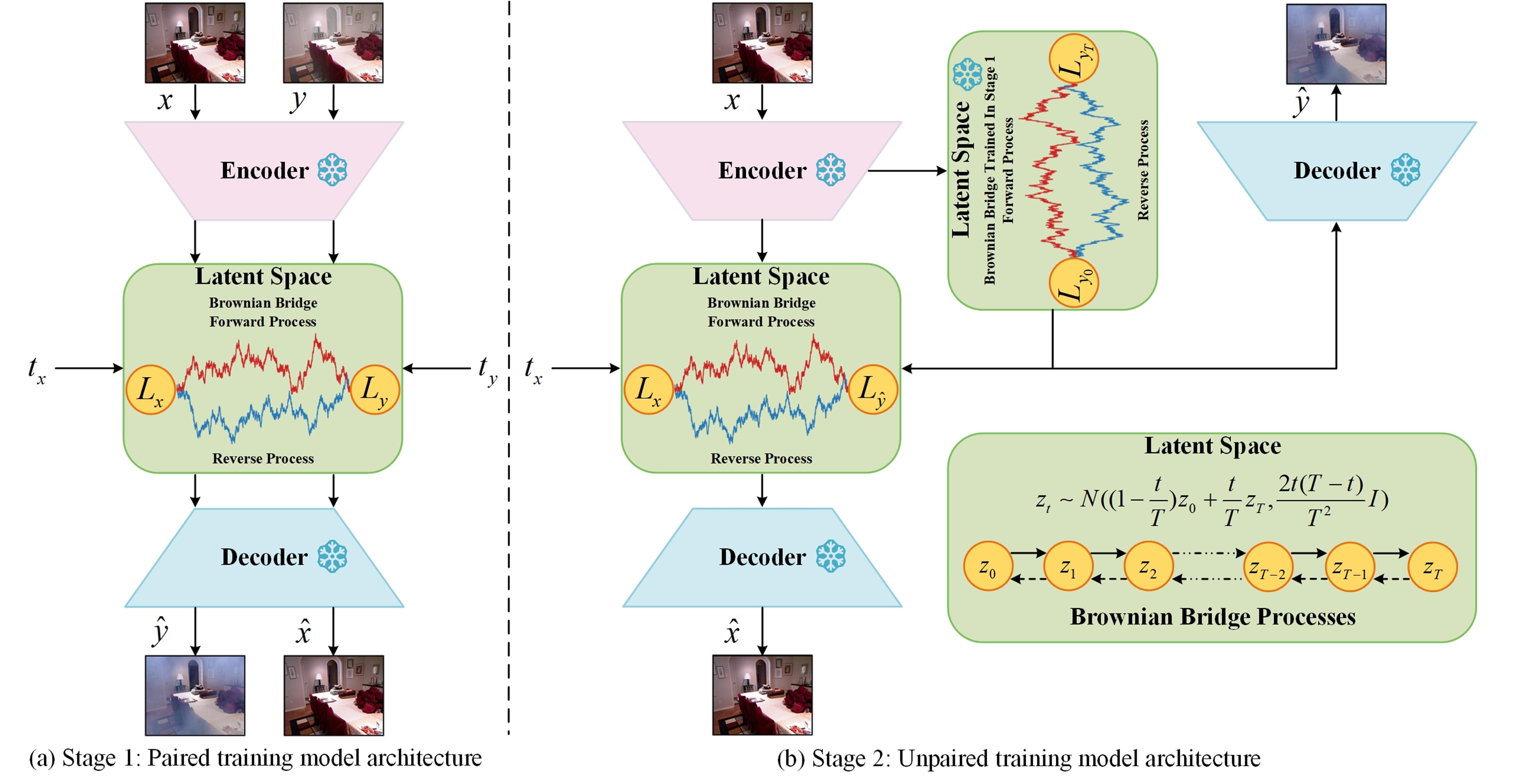}
\caption{\textbf{Architecture of our proposed EM-B\textsuperscript{3}DM.} (a) In the first stage, paired data is utilized for training, with the joint distribution optimized using the EM algorithm. (b) In the second stage, our EM-B\textsuperscript{3}DM undergoes semi-supervised training leveraging the joint distribution learned in the first stage, with all paired data  abandoned.}
\label{fig:1}
\end{figure*}


\section{METHOD}
Inspired by diffusion models, we introduce a novel semi-supervised image dehazing framework that integrates the Expectation Maximization algorithm with a stochastic Brownian Bridge diffusion process, termed EM-B\textsuperscript{3}DM, which exhibits greater stability and maintains good convergence during training. The pipeline of the proposed method is shown in Fig.~\ref{fig:1} and is detailed as follows. Formally, suppose we have clear images $x$ and hazy images $y$ sampled from a joint distribution $q(x,y)$. To alleviate the computational overhead in training, we move the diffusion process in the latent space of VQGAN \cite{c:50}. We add the corresponding annotations in the pipeline Fig.~\ref{fig:1} to mark the fixed parts clearly. For simplicity, we still use $x$ and $y$ to denote the corresponding latent features $(x:=L_{x},y:=L_{y})$.
\subsection{Model Design}

\begin{figure*}[t]
\centering
\includegraphics[width=1\textwidth]{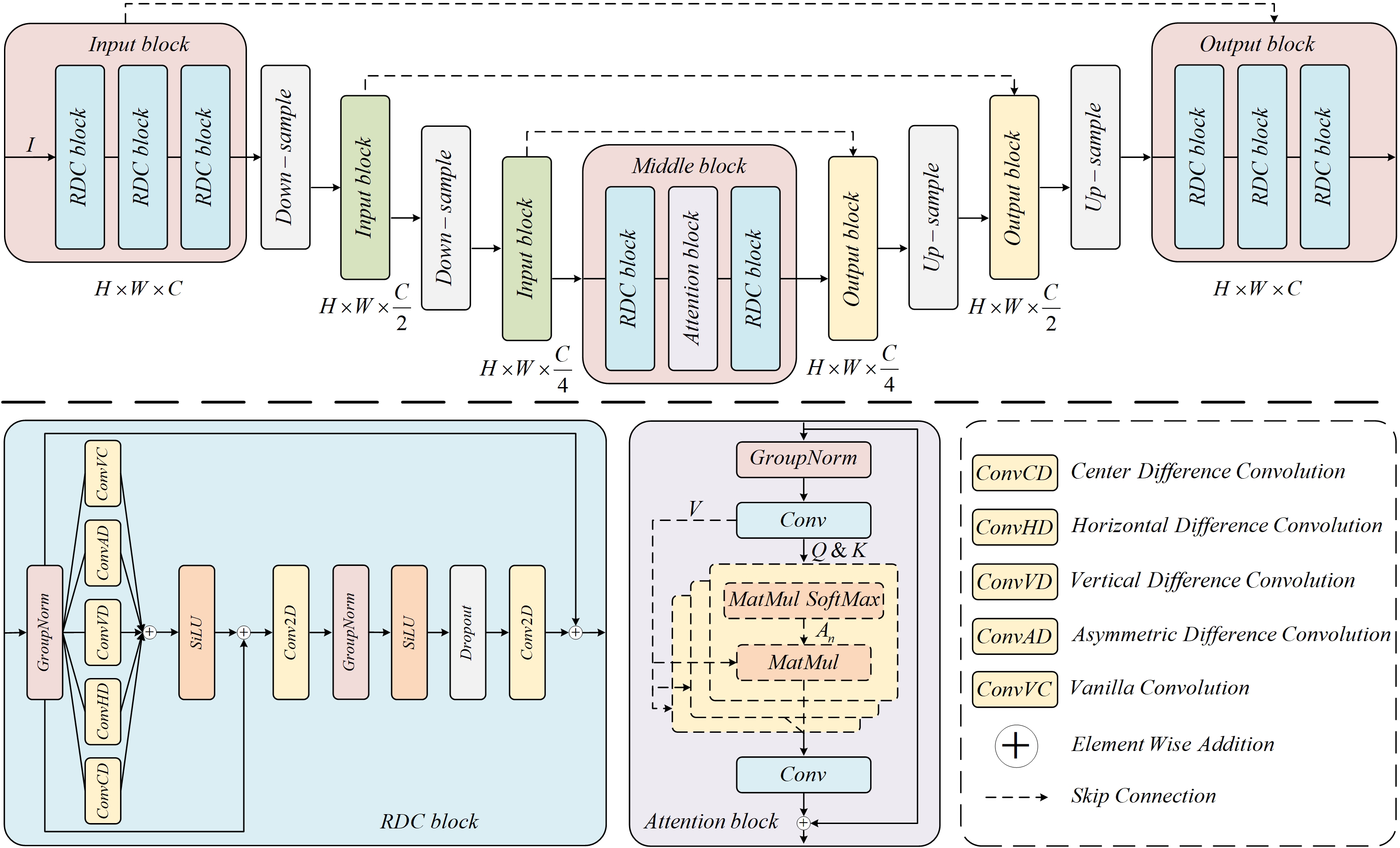}
\caption{\textbf{Overview of the noise predictor network architecture.} The network employs the encoder-decoder structure of U-Net, combined with skip connections to better capture the spatial information of the image.}
\label{fig:3}
\end{figure*}
  Building upon earlier diffusion models, such as the Denoising Diffusion Probabilistic Model (DDPM), aim to ensure that $p(x)$ closely approximates $q(x)$ by maximizing the following log-likelihood function: 
\begin{equation}
  \theta =\underset{\theta}{argmax}\int q(x)logp(x)dx ,
  \label{eq:6}
\end{equation}
which is equivalent to minimizing $KL(q(x)||p(x))$:
\begin{equation}
  KL(q(x)||p(x)) = \int q(x)log\frac{q(x)}{p(x)}dx.
  \label{eq:7}
\end{equation}
We aim to design a diffusion model for self-supervised image dehazing by utilizing the randomness inherent in the diffusion process and the EM algorithm to decouple the joint distribution $q(x,y)$. This is achieved by reformulating the objective through variational inference to minimize the KL-divergence of the joint distribution $KL(q(x,y)||p(x,y))$ instead of the marginal KL-divergence $KL(q(x)||p(x))$. We have
\begin{equation}
\begin{aligned}
  &KL(q(x,y)||p(x,y)) \\
  &=KL(q(x)||p(x))+\int q(x)KL(q(y|x)||p(y|x))dx \\
  &\ge KL(q(x)||p(x)),
\end{aligned}
  \label{eq:8}
\end{equation}
indicating that $KL(q(x,y)||p(x,y))$ provides an upper bound on $KL(q(x)||p(x))$. If it works, we have $p(x,y)\to q(x,y)$. Considering that image dehazing and hazing are inherently conditional generation tasks, it becomes superfluous to model the complete joint distribution directly. We let $q(x,y)=q_{\theta}(y|x)q(x)$ and $p(x,y)=p_{\theta}(x|y)p(y)$, while $q_{\theta}(y|x)$, $p_{\theta}(x|y)$ are Gaussian distributions modeled using a unified Brownian Bridge Diffusion Model, with unknown parameters. Consequently, we can express the KL divergence in \cref{eq:8} as:
\begin{equation}
E_{x\sim q(x)}[\int q_{\theta}(y|x)log\frac{q_{\theta}(y|x)}{p_{\theta}(x|y)p(y)}dy]+E[logq(x)],
  \label{eq:9}
\end{equation}
while $q(x)$, $p(y)$ do not contain any parameters and can be treated as constants. Both $q_{\theta}(y|x)$ and $p_{\theta}(x|y)$ are unknown, and by assuming $q_{\theta}(y|x)$ as a constant in accordance with the EM algorithm. At the \( r \)-th iteration, we can optimize $p_{\theta^{(r)}}(x|y)$ by maximizing:
\begin{equation}
\begin{aligned}
\underset{p_{\theta}(x|y)}{argmax}  E_{x\sim q(x)}[\int q_{\theta^{(r-1)}}(y|x)log[p_{\theta}(x|y)]dy].
  \label{eq:10}
\end{aligned}
\end{equation}
Subsequently, we assume $p_{\theta^{(r)}}(x|y)$ to be known and proceed to optimize $q_{\theta^{(r)}}(y|x)$ using the following equation:
\begin{equation}
\begin{aligned}
\underset{q_{\theta}(y|x)}{argmin}  E_{x\sim q(x)}[\int q_{\theta}(y|x)log\frac{q_{\theta}(y|x)}{p(y|x)p(x)} dy],
  \label{eq:11}
\end{aligned}
\end{equation}
where $p(y|x)=\frac{p_{\theta^{(r)}}(x|y)p(y)}{p(x)}, p(x)=\int p_{\theta^{(r)}}(x|y)p(y)dy$.

\subsection{Brownian Bridge Processes}
\label{sec:Brownian}
To better align with the dynamic nature of the diffusion process and the mathematical formulation of the Brownian Bridge, we introduce $z_{0}$ and $z_{T}$ to denote the starting and ending states of the Brownian Bridge process. This notation explicitly captures the dynamic transition from $z_{0}$ (input latent feature) to $z_{T}$ (target latent feature), emphasizing the constraints of the starting and ending points. This shift from static feature representations ($x$ and $y$) to dynamic states ($z$) allows for a clearer and more precise modeling of the diffusion process.
\subsubsection{Forward Process.}
The forward process starts from the initial state distribution $z_{0}$ and gradually injects noise over T steps, ultimately reaching the final state distribution $z_{T}$. In the case of $q_{\theta}(y|x)$, the starting state is $z_{0}=x$ and the ending state is $z_{T}=y$. The forward diffusion process of Brownian Bridge can be formalized by a Markov chain:
\begin{equation}
\begin{aligned}
q^{BB}(z_{t}|z_{0},z_{T})=\mathcal{N}(z_{t};&(1-m_{t})z_{0}+m_{t}z_{T},\delta _{t}I), \\
m_{t}=\frac{t}{T}, t&=1,2,\cdots,T, 
  \label{eq:12}
\end{aligned}
\end{equation}
where $T$ is the total steps of the diffusion process, $\delta _{t}$ is the variance, $I$ is the identity matrix. We take the noise variance of Brownian Bridge process $\delta _{t}=2s(m_{t}-m_{t}^{2})$ as mentioned in \cite{c:37}. The transition probability $q_{BB}(z_{t}|z_{t-1},z_{T})$ can be derived from \cref{eq:12}
\begin{equation}
\begin{aligned}
q^{BB}(z_{t}|z_{t-1},z_{T})=\mathcal{N}(z_{t};\frac{1-m_{t}}{1-m_{t-1}z_{t-1}}\\
+(m_{t}-\frac{1-m_{t}}{1-m_{t-1}}m_{t-1})z_{T},\delta _{t|t-1}I ), 
  \label{eq:13}
\end{aligned}
\end{equation}
where $\delta _{t|t-1}$ is defined as follows:
\begin{equation}
\begin{aligned}
\delta _{t|t-1}= \delta _{t}-\delta _{t-1}\frac{(1-m_{t})^{2}}{(1-m_{t-1})^{2}}.
  \label{eq:14}
\end{aligned}
\end{equation}

\subsubsection{Reverse Process.} 

The reverse process aims to estimate the posterior distribution $p_{BB}(z_{0}|z_{T})$ via the following formulation:
\begin{equation}
\begin{aligned}
p^{BB}(z_{0}|z_{T})=\int p(z_{T})\prod_{t=1}^{T} p^{BB}_{\theta}(z_{t-1}|z_{t})dz_{1:T},
  \label{eq:15}
\end{aligned}
\end{equation}
where $p_{\theta}^{BB}$ is the inverse transition kernel from $z_{t}$ to $z_{t-1}$ with a learnable parameter $\theta$.
\begin{equation}
\begin{aligned}
p_{\theta}^{BB}(z_{t-1}|z_{t},z_{T})= \mathcal{N}(z_{t-1};\mu_{\theta}, \widetilde{\delta_{t}}I ),
  \label{eq:16}
\end{aligned}
\end{equation}
where $\mu_{\theta}$ is the predicted mean value of the noise, and $\widetilde{\delta_{t}}$ is the variance of noise at each step.

\subsection{Training Objective}
\label{sec:Objective}
We reformulate the optimization of $p_{\theta}(x|y)$ and $q_{\theta}(y|x)$ as an optimization problem based on the Brownian Bridge diffusion model. Accordingly, we can express the joint probability in terms of a path-dependent optimization problem. Removing the constant term, \cref{eq:10} and \cref{eq:11} can be unified and transformed according to \cref{eq:15}:
\begin{equation}
\begin{aligned}
\underset{\theta^{(r)}}{argmax}  E[\int p(z_{T})\prod_{t=1}^{T} p^{BB}_{\theta^{(r)}}(z_{t-1}|z_{t})dz_{1:T}]. 
  \label{eq:17}
\end{aligned}
\end{equation}
The optimization for $\theta^{(r)}$ is achieved by minimizing the negative evidence lower bound, namely,
\begin{equation}
\begin{aligned}
\sum_{t=2}^{T} D_{KL}(q^{BB}(z_{t-1}|z_{t},z_{0},z_{T})||p^{BB}_{\theta^{(r)}}(z_{t-1}|z_{t},z_{T})), 
  \label{eq:18}
\end{aligned}
\end{equation}
More mathematical details can be found in supplementary. \textbf{Combining} \cref{eq:12} and \cref{eq:13}, we obtain a tractable form of the target distribution $q^{BB}(z_{t-1}|z_{t},z_{0},z_{t})$, which can be explicitly written as follows:
\begin{equation}
\begin{aligned}
q^{BB}(z_{t-1}|z_{t},z_{0},z_{T})&=\mathcal{N}(z_{t-1};c_{t}z_{t}+c_{T}z_{T}\\
&+c_{\epsilon}\epsilon_{\theta^{(r)}}, \widetilde{\delta_{t}}I ) 
  \label{eq:19}
\end{aligned}
\end{equation}
where $\epsilon_{\theta^{(r)}}$ is a deep neural network with parameter $\theta^{(r)}$, aiming to predict $z_{0}$.
\begin{equation}
\begin{aligned}
c_{t}&=\frac{\delta _{t-1}}{\delta _{t}}\frac{1-m_{t}}{1-m_{t-1}}+(1-m_{t-1})\frac{\delta _{t|t-1}}{\delta _{t}}, \\
c_{T}&=m_{t-1}-m_{t}\frac{1-m_{t}}{1-m_{t-1}}\frac{\delta _{t-1}}{\delta _{t}},\\
c_{\epsilon }&= (1-m_{t-1})\frac{\delta _{t|t-1}}{\delta _{t}}, \widetilde{\delta_{t}}=\frac{\delta_{t|t-1}\cdot \delta _{t-1}}{\delta _{t}}.
  \label{eq:20}
\end{aligned}
\end{equation}
Based on \cref{eq:16} and \cref{eq:19},  we simplify the objective function as follows,
\begin{equation}
\begin{aligned}
\underset{\theta ^{(r)}}{min} \sum _{t}||\epsilon_{\theta ^{(r)}}(z_{t},z_{T},t)-z_{0} ||_{1}.
  \label{eq:21}
\end{aligned}
\end{equation}
\subsection{Network Architecture}
Our network architecture is shown in Fig.~\ref{fig:3}, we implement modifications to the U-Net structure in DDPM \cite{c:21} and introduce the RDC block. Leveraging the properties of difference convolutions in capturing gradient-level information \cite{c:38}, we deploy four difference convolutions (ConvDC, ConvAD, ConvHD, ConvVD) alongside one vanilla convolution for feature extraction. By exploiting reparameterization and the additivity of convolution layers, we use the vanilla convolution to capture intensity-level details, while the difference convolutions enhance gradient-level features. 

\subsection{Semi-supervised Training Process}
As shown in \cref{alg:training}, our training process is divided into two stages. In the first stage, the EM algorithm is used to optimize the joint distribution of paired hazy and clear images. In the second stage, we fix $\epsilon_{\theta_{1}}$ obtained from the first stage and use it for unpaired training. Based on \cref{sec:Brownian} and \cref{sec:Objective} optimizing the conditional distributions $p(x|y)$ and $q(y|x)$ is equivalent to estimating the conditional expectation of the noise injected into $z_{t}$. The two conditional distributions are modeled using the Brownian Bridge diffusion process, differing only in their respective starting and ending states. Inspired by a unified framework, we express conditional expectations in the general form $E[\epsilon_{x},\epsilon_{y}|z_{t_{x}},z_{t_{y}}]$, where $t_{x}$ and $t_{y}$ are two timesteps that can be different.

\begin{algorithm}[t]
\caption{Two-Stage Training Process}
\label{alg:training}
\begin{algorithmic}[1] 
\STATE \textbf{Stage 1: Paired Training}
\STATE  \textbf{Notation:} 
\STATE \quad $x:=L_{x}, y:=L_{y}$
\STATE \quad $z_{t_x}, z_{t_y}$ are the diffused latent variables at timesteps $t_x$, $t_y$
\STATE \quad $z_t$ is the transition from $z_0$ to $z_T$ in the single-bridge setting
\STATE \textbf{repeat:}
    \STATE \quad $x, y \sim q(x_0, y_0)$
    \STATE \quad $t_{x}, t_{y} \sim Uniform(1, ...,T)$
    \STATE \quad Gaussian noise $\epsilon_{t_{x}}, \epsilon_{t_{y}} \sim \mathcal{N}(0, I)$
    \STATE \quad $z_{t_{x}} = (1-m_{t_{x}})x + m_{t_{x}}y + \sqrt{\delta_{t_{x}}}\epsilon_{t_{x}}$
    \STATE \quad $z_{t_{y}} = (1-m_{t_{y}})y + m_{t_{y}}x + \sqrt{\delta_{t_{y}}}\epsilon_{t_{y}}$
    \STATE \quad Take gradient step on:
    \STATE \quad \quad $\nabla_{\theta_{1}} ||\epsilon_{\theta_{1}}(z_{t_{x}}, z_{t_{y}}, t_{x}, t_{y}) - (x, y)||_{1}$
\STATE \textbf{until} convergence

\STATE \textbf{Stage 2: Unpaired Training}
\STATE \textbf{repeat:}
    \STATE \quad $z_{T}=x \sim q(x)$
    \STATE \quad $\widehat{y} = \epsilon_{\theta_{1}}(x,z_{T},0,T)$
    \STATE \quad $t \sim Uniform(1, ...,T), \epsilon_{t}\sim \mathcal{N}(0, I)$
    \STATE \quad $z_{t} = (1-m_{t})x + m_{t}\widehat{y} + \sqrt{\delta_{t}}\epsilon_{t},z_{T}=\widehat{y}$
    \STATE \quad Take gradient step on:
    \STATE \quad \quad $\nabla_{\theta} ||\epsilon_{\theta}(z_{t},z_{T},t)-x||_{1}$
\STATE \textbf{until} convergence
\end{algorithmic}
\end{algorithm}

\begin{algorithm}[t]
\caption{Sampling}
\label{alg:Sampling}
\begin{algorithmic}[1] 
\STATE $z_{T}=y \sim q(y)$
\FOR{$t = T, ...,1$}
    \STATE $\epsilon_{t}\sim \mathcal{N}(0, I)$ if $t>1$, else $\epsilon_{t}=0$
    \STATE $z_{t-1} = (1-m_{t})\epsilon_{\theta}(z_{t},z_{T},t) + m_{t}y + \sqrt{\delta_{t}}\epsilon_{t}$
\ENDFOR
\STATE \textbf{return} $z_{0}$
\end{algorithmic}
\end{algorithm}

\begin{table*}[t]
\centering
\begin{tabular}{ccccccccc}
    \toprule
    \multicolumn{2}{c}{\multirow{2}{*}{\textbf{Methods}}} & \multicolumn{2}{c}{SOTS-indoor \cite{c:39}} & \multicolumn{2}{c}{SOTS-outdoor \cite{c:39}} & \multicolumn{2}{c}{NH-HAZE 2 \cite{c:40}}  \\
    \cmidrule(lr){3-4} \cmidrule(lr){5-6} \cmidrule{7-8} 
    & & PSNR$\uparrow$ & SSIM$\uparrow$ & PSNR$\uparrow$ & SSIM$\uparrow$ & PSNR$\uparrow$ & SSIM$\uparrow$  \\
    \midrule
    \midrule
    
    \multirow{4}{*}{\textbf{Supervised}} & DehazeNet \cite{c:6} & 19.82 & 0.818 & 24.75 & 0.927 & 10.62 & 0.521  \\
    &AOD-Net \cite{c:7} & 20.51 & 0.816 & 24.14 & 0.920 & 12.33 & 0.631  \\
    &MSCNN \cite{c:42} & 19.84 & 0.833 & 14.62 & 0.908 & 11.74 & 0.566  \\
    &GDN \cite{c:43} & \textbf{32.16} & \textbf{0.983} & 17.69 & 0.841 & 12.04 & 0.557 \\
    \midrule
    \midrule
    \multirow{3}{*}{\textbf{Semi-supervised}}
    & DA \cite{c:56}  & 27.79 & 0.918 & 22.88 &  0.872 & - & - \\
    & PSD \cite{c:57} & 14.73 &  0.685 &  15.13 &  0.723 & 12.23 & 0.670 \\
    & DTS \cite{c:58} & 28.32 & 0.928 & \underline{26.70} & 0.947 & - & - \\
    \cmidrule{2-8}
    &~ \textbf{Ours}  &~ \underline{28.85} &~ 0.924 &~ \textbf{28.96} &~ 0.909 &~ \textbf{19.37} &~ 0.703  \\
    \bottomrule
\end{tabular}
\caption{Quantitative comparisons of different dehazing methods across datasets.}
\label{table:dehazing_comparison_with_overhead}
\end{table*}

\section{Experiments}
\subsection{Experimental Configuration}
\subsubsection{Datasets.} 

We train and evaluate the proposed method on publicly available datasets SOTS \cite{c:39} and NH-HAZE 2 \cite{c:40}. The ratio of paired data to unpaired data was 1:1. The SOTS synthetic dataset contains indoor and outdoor subsets. In the indoor dataset, there are 500 pairs of hazy and haze-free images. We randomly split the dataset into 450 training pairs and 50 test pairs. Among the 450 training pairs, 225 pairs were randomly selected as the paired dataset, and the remaining 225 hazy images were used as the unpaired  dataset. The processing of the outdoor dataset followed the same approach as the indoor dataset. For NH-HAZE2, we also divided the training set into paired and unpaired data with a 1:1 ratio. Due to the architectural constraints of UNet, we adopted a patch-based training strategy, where 256×256 patches were randomly cropped from high-resolution images. This approach not only ensures compatibility with the model but also acts as an effective form of data augmentation, significantly increasing the diversity of training samples. Although we randomly cropped 256×256 images for training due to UNet limitations, for fair comparison, we adopted a sliding window \cite{c:60} to process images at their original resolution during evaluation. Furthermore, for a 640×480 image, we can obtain $(640-256+1)(480-256+1) = 86,625$ possible images, enhancing data efficiency and performance on small datasets.

\subsubsection{Compared Methods.} 

We evaluate the performance of EM-B\textsuperscript{3}DM against various state-of-the-art dehazing methods, including the prior-based method (DCP \cite{c:44}), supervised methods trained on the entire training set (DehazeNet \cite{c:6}, AOD-Net \cite{c:7}, MSCNN \cite{c:42}, GDN \cite{c:43}), semi-supervised methods (DA \cite{c:56}, PSD \cite{c:57}, DTS \cite{c:58}) that use the same training set as DTS. For fair comparison, we utilize the official code provided by the respective authors and retrained these methods under the same experimental settings. For methods that could not be retrained, we use the given models for testing purposes.

\subsubsection{Training Details.}
The proposed  is implemented using PyTorch 1.12.1 and trained on a PC Intel(R) Core(TM) i5-13600K CPU @ 5.10GHz and an NVIDIA GeForce RTX 4090 GPU. The  framework consists of two components: a pretrained VQGAN model and the proposed Brownian Bridge diffusion model. We utilize the same pretrained VQGAN model as employed in the Latent Diffusion Model \cite{c:49}. Timesteps is set to 1000 during the training phase, while 200 sampling steps are used during inference to balance sample quality and efficiency. The optimizer used is Adam, with the default PyTorch settings and a fixed learning rate of 5e-5. Our training strategy is structured in two stages: the first stage involves a training process using limited paired data, while the second stage exclusively utilizes a large amount of unpaired data for training. It is noteworthy that the second stage does not fine-tune the model from the first stage. In fact, it is trained entirely based on the unpaired data.

\subsection{Performance Evaluation}

\subsubsection{Quantitative Analysis.} 
As shown in Table~\ref{table:dehazing_comparison_with_overhead}, we present a detailed quantitative comparison of various state-of-the-art dehazing methods across multiple datasets, including SOTS-indoor, SOTS-outdoor, and NH-HAZE 2. Our EM-B\textsuperscript{3}DM achieves the highest PSNR on the SOTS-Outdoor (28.96 dB) and NH-HAZE 2 (19.37 dB) datasets, along with competitive SSIM values. The results above demonstrate that our EM-B\textsuperscript{3}DM inherits the advantages of diffusion models in generating realistic images. Additionally, we employed non-reference evaluation metrics NIQE and MUSIQ to assess different methods, as shown in Table~\ref{tab:nm}.

\begin{figure*}[t]
\centering
\includegraphics[width=1\textwidth]{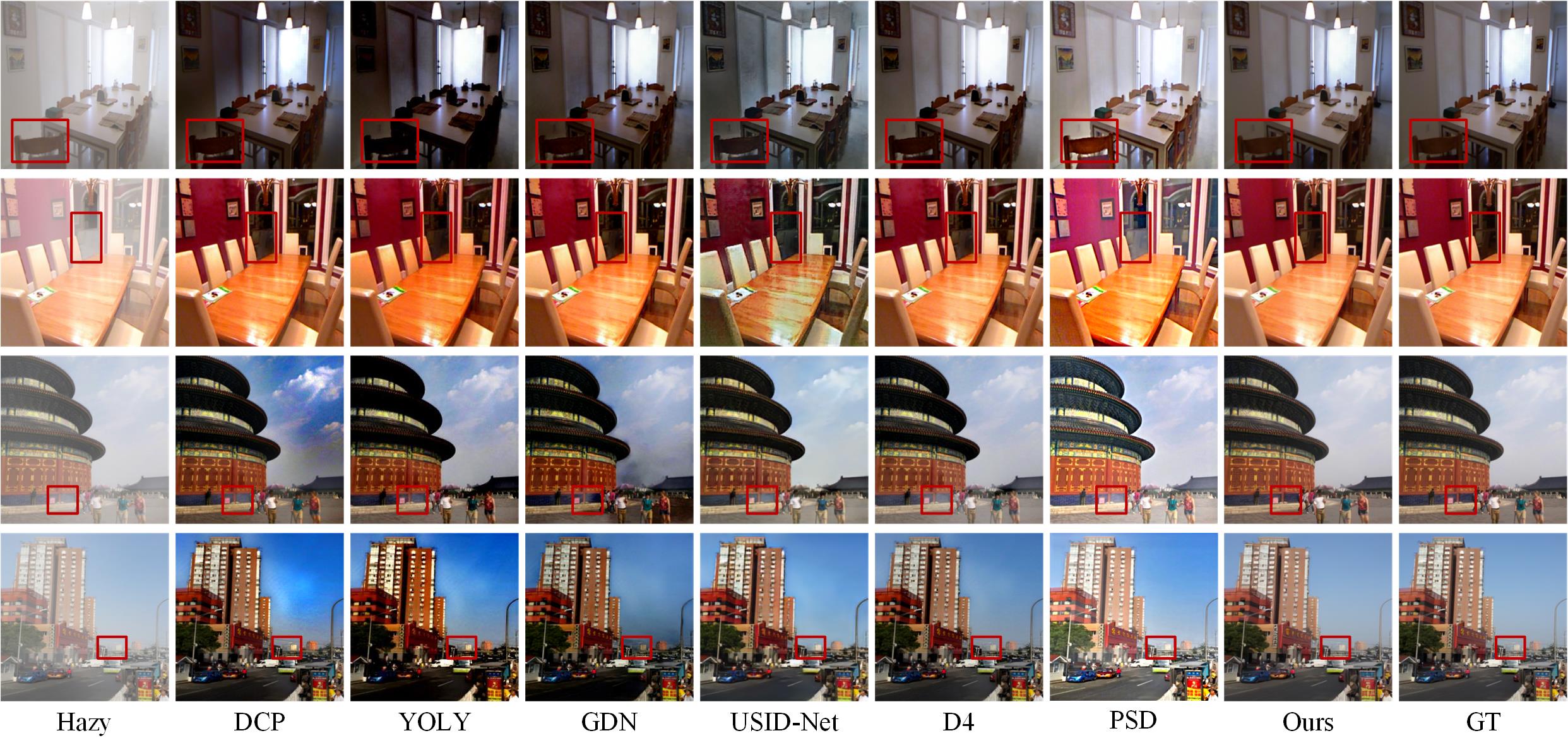}
\caption{Visual comparison of various dehazing methods on SOTS-indoor and SOTS-outdoor. Please zoomin on screen for a better view. }
\label{fig:4}
\end{figure*}

\begin{table}[H]
  \centering
  \small
  \begin{tabular}{ccccc}
    \toprule
    \multicolumn{1}{c}{\multirow{2}{*}{\textbf{Methods}}}& \multicolumn{2}{c}{SOTS-indoor \cite{c:39}} & \multicolumn{2}{c}{SOTS-outdoor \cite{c:39}} \\
    \cmidrule(lr){2-3} \cmidrule(lr){4-5}
    & NIQE $\downarrow$ & MUSIQ $\uparrow$ &  NIQE $\downarrow$ & MUSIQ $\uparrow$ \\
    \midrule
    \midrule
    DA&  5.415& 44.549 &   5.875 &  47.342 \\
    PSD&  4.850 & 48.645 &   5.445 &58.915 \\
    DTS&   5.172&    51.206     &  5.905 &  56.071 \\
    \midrule
    ~ \textbf{Ours} & ~ \textbf{4.632} &  ~ \textbf{56.720} &  ~ \textbf{4.594} & ~ \textbf{61.182} \\
    \bottomrule
  \end{tabular}
  \caption{The evaluation results of the NIQE and MUSIQ.}
  \label{tab:nm}
\end{table}
\vspace{-1.5em}
\begin{table}[H]
  \centering
  \small
  \begin{tabular}{ccccc}
      \toprule
      Methods& \multicolumn{2}{c}{O-Haze} & \multicolumn{2}{c}{Dense-Haze} \\
      \cmidrule(lr){2-3} \cmidrule(lr){4-5}
      & PSNR $\uparrow$ & SSIM $\uparrow$ &  PSNR $\uparrow$ & SSIM $\uparrow$ \\
      \midrule
      \midrule
      D4(base) &  16.922 & \underline{0.607} &  11.412 &  \underline{0.385} \\
      DDPM&  15.163 & 0.492 & 10.147 &  0.295 \\
      AutoDIR&  16.517 & 0.615 &  11.642 &  0.332 \\
      WeatherDiff &  17.195 & 0.603 &  12.059 &  0.318 \\
      Ours&  \textbf{18.334} & \textbf{0.657} &  \textbf{13.826} & \textbf{0.441} \\
      Ours(Semi)&  \underline{17.645} & 0.591 &  \underline{13.277} & 0.361 \\
      \bottomrule
    \end{tabular}
  \caption{Quantitative comparison on two real-world datasets}
  \label{tab:nm}
\end{table}
\vspace{-1em}
\subsubsection{Compared to existing diffusion-based methods}
Our EM-B\textsuperscript{3}DM has several unique advantages over traditional diffusion models. It enhances the dynamic modeling of both forward and reverse diffusion processes using the Brownian Bridge, making the transition between hazy and clean images more structured. This bidirectional modeling ability allows for more precise learning of the data distribution, thereby improving the consistency of dehazing results. Through this bidirectional modeling, the model can generate pseudo-labels for self-supervised training, significantly reducing the reliance on paired data.
We provide additional experimental results on two real-world datasets in Table~\ref{tab:nm}. Compared to existing diffusion-based methods such as DDPM, AutoDIR, and WeatherDiff, which rely on fully paired training data, our model adopts a semi-supervised approach with a 1:1 split of the available training data into paired and unpaired samples. While DDPM, AutoDIR, and WeatherDiff require extensive paired datasets to achieve strong performance, our method demonstrates competitive dehazing capabilities while significantly reducing the dependence on large amounts of paired data, highlighting its practicality and effectiveness in scenarios with limited supervision.

DDPM, AutoDIR, WeatherDiff and EM-B\textsuperscript{3}DM both incorporate the basic idea of DDIM and utilize a non-Markovian process to accelerate inference with 200 timesteps. DDPM, AutoDIR and WeatherDiff fail to generate satisfactory results due to its mechanism of integrating conditional input into the diffusion model. In contrast, EM-B\textsuperscript{3}DM directly learns a diffusion process between these two domains, avoiding the reliance on conditional information. 

\subsubsection{Qualitative Analysis.} 

Fig.~\ref{fig:4} presents a visual comparison between our proposed EM-B\textsuperscript{3}DM and previous state-of-the-art methods on the SOTS-indoor and SOTS-outdoor datasets. Methods such as YOLY, GDN, and USID-Net result in darkened areas and blurred contours, while PSD introduces color distortions. In contrast, our proposed EM-B\textsuperscript{3}DM recovers sharper contours and edges, with fewer haze remnants in the output. DCP and YOLY occasionally fail to handle sky regions, leading to significant color shifts and artifacts in the processed images. Our EM-B\textsuperscript{3}DM, however, produces results that are closer to the ground truth, providing visually more appealing dehazing effects.

\subsection{Ablation Study}
\subsubsection{Sampling Steps.} 
EM-B\textsuperscript{3}DM incorporates DDIM \cite{c:51}, leveraging a non-Markovian process to accelerate inference. We investigate the impact of sampling steps in the reverse diffusion process on the performance of EM-B\textsuperscript{3}DM by conducting evaluations across different sampling steps on the SOTS-indoor dataset. The quantitative scores for the image dehazing task are reported in Table~\ref{tab:timesteps}. The results show that when the number of sampling steps is relatively small (fewer than 200), image quality improves rapidly as the step count increases. When the number of steps exceeds 200, the PSNR and SSIM metrics display only slight improvements.
\begin{table}[htbp]
  \centering
  \begin{tabular}{cccc}
    \toprule
    \multicolumn{2}{c}{Configurations} & \multicolumn{2}{c}{Metrics} \\
    \cmidrule(lr){1-2} \cmidrule(lr){3-4}
    Factor $s$ & Sampling Steps $T$ & PSNR $\uparrow$ & SSIM $\uparrow$ \\   
    \midrule
    \midrule
    \multirow{5}{*}{1.0} &15   & 27.54 & 0.908 \\
                        &50   & 27.93 & 0.912 \\
                        &100  & 28.42 & 0.919 \\
                        &200  & 28.85 & 0.924 \\
                        &1000 & 28.87 & 0.925\\
    \midrule
    \midrule
    0.5& \multirow{4}{*}{200} &  27.42 & 0.902 \\
    1.0 &                    &  28.85 & 0.924 \\
    2.0 &                    &  27.74 & 0.893 \\
    4.0 &                    &  26.49 & 0.887 \\
    \bottomrule
  \end{tabular}
  \caption{Ablation analysis on model configurations.}
  \label{tab:timesteps}
\end{table}

\subsubsection{Hyper-paramete $s$.} 
The variance of the Brownian Bridge characterizes its temporal uncertainty and fluctuation. The dynamic evolution of this variance is leveraged to control the randomness inherent in the generative process. The Brownian Bridge diffusion process stabilizes at the endpoints while maintaining heightened stochasticity in the intermediate stages, ensuring sufficient diversity and flexibility during generation. This variance modulation is critical for facilitating a smooth transition between the initial and target distributions. As shown in \cref{eq:12}, the quality of the generated images can be regulated by scaling the maximum variance of the Brownian Bridge at $t=\frac{T}{2}$ with a factor $s$. In this section, we experiment with $s\in \left \{ 0.5,1,2,4 \right \}$ to assess its impact on performance of EM-B\textsuperscript{3}DM. The quantitative results are presented in Table~\ref{tab:timesteps}. With increasing $s$, the quality and fidelity of the generated images degrade. If the original variance formulation of the Brownian Bridge is used without scaling, EM-B\textsuperscript{3}DM fails to generate plausible samples due to the excessively large maximum variance.

\begin{table}[H]
  \centering
  \small
  \begin{tabular}{c|ccccc}
    \toprule
    Design & Base  & RDC & RDC & RDC & RDC \\
    \midrule
    \midrule
    Vanilla Conv & \CheckmarkBold & \CheckmarkBold & \CheckmarkBold & \CheckmarkBold & \CheckmarkBold\\
    $+$ ConvCD &                & \CheckmarkBold & \CheckmarkBold & \CheckmarkBold & \CheckmarkBold\\
    $+$ ConvAD &                &                & \CheckmarkBold & \CheckmarkBold & \CheckmarkBold\\
    $+$ ConvHD &                &                &                & \CheckmarkBold & \CheckmarkBold\\
    $+$ ConvVD &                &                &                &  & \CheckmarkBold\\
    \midrule
    PSNR $\uparrow$ & 27.60& 27.33& 27.65& 28.42 & \textbf{28.85} \\
    SSIM $\uparrow$ & 0.903& 0.907& 0.912& 0.918 & \textbf{0.924} \\
    \bottomrule
  \end{tabular}
  \caption{Ablation on RDC block.}
  \label{tab:RDC}
\end{table}

\subsubsection{Effectiveness of RDC block.} 

\begin{figure}[htbp]
\centering
\includegraphics[width=\columnwidth]{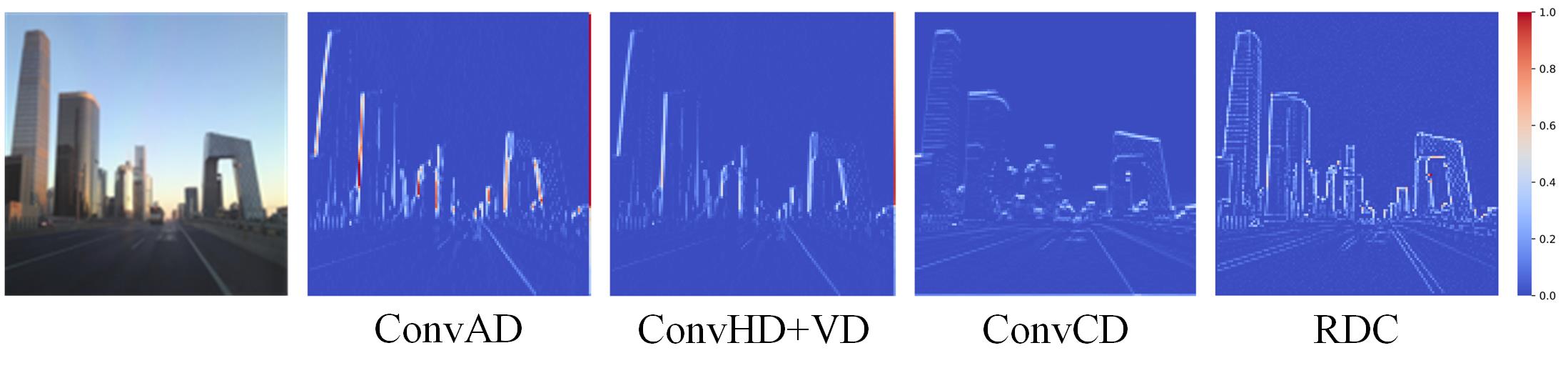}
\caption{Visualization of Different Differential Convolution Features.}
\label{fig:5}
\end{figure}

To demonstrate the effectiveness of the proposed RDC block, we conduct an ablation study with 500k training iterations to assess its contribution. Specifically, we construct the ResBlock in UNet \cite{c:21} as the baseline and ablate the five parallel convolution layers, and the results are summarized in Table~\ref{tab:RDC}. As shown by the results, the PSNR performance steadily improves from 27.60 dB to 28.85 dB, with SSIM exhibiting a similar trend. As shown in Fig.~\ref{fig:5}, ConvAD focuses on regions with significant angular variation, while ConvHD and ConvVD primarily capture gradient changes in the horizontal and vertical directions, highlighting edge or texture variations. ConvCD enhances detail and texture variation. Our RDC effectively provides a comprehensive representation that captures diverse textures, edges, and fine details across the image.

\subsubsection{Effectiveness of Stage 1.} 
\begin{figure}[htbp]
\centering
\includegraphics[width=\columnwidth]{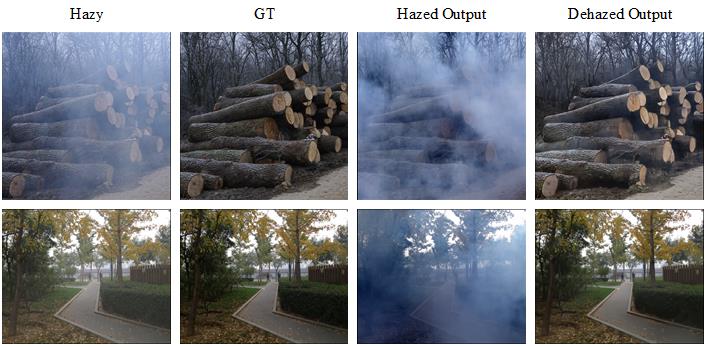}
\caption{Some dehazing results of the Stage 1 pre-trained model.}
\label{fig:6}
\end{figure}

The objective function of Stage 1 in Algorithm 1 unifies the prediction of both conditional distributions by learning the noise in the perturbed data, allowing a single network to learn both conditional distributions simultaneously. This joint modeling capability allows the network to better capture the bidirectional transformation between hazy and clean images, ensuring a more accurate representation of the underlying distribution. Furthermore, the ability to learn both conditional distributions in Stage 1 plays a crucial role in the subsequent training phase. Specifically, in Stage 2, we leverage the pre-trained Stage 1 model to generate pseudo-labels for self-supervised dehazing. By synthesizing paired clean-hazy data through the learned bidirectional mapping, the Stage 1 model effectively provides high-quality supervision signals, reducing the reliance on manually annotated datasets. To validate the effectiveness of Stage 1, we present some dehazing results from the pre-trained model in Fig.~\ref{fig:6}, demonstrating its capability to recover clean images from hazy inputs with high fidelity.

\subsubsection{Contribution of the Framework and Novelty of Our Method.} The theoretical foundation of BBDM \cite{c:59} is based on the KL divergence of marginal distributions $KL(q(x)||p(x))$, focusing on unidirectional image translation, such as translating from domain A to domain B. In contrast, EM-B\textsuperscript{3}DM utilizes the KL divergence of joint distributions $KL(q(x,y)||p(x,y))$, which is theoretically decoupled using the EM algorithm (Sec. 3.1). In Stage 1, EM-B\textsuperscript{3}DM extends BBDM to unify a single model to perform bi-directional generation simultaneously with limited paired data. In the second stage, EM-B\textsuperscript{3}DM successfully implements semi-supervised learning by generating pseudo-paired data for image dehazing.

\section{Conclusion}

In this paper, we propose EM-B\textsuperscript{3}DM, a novel semi-supervised image dehazing framework that utilizes the Expectation-Maximization (EM) algorithm and bidirectional Brownian Bridge diffusion models to model the joint distribution $q(x, y)$. The first training stage uses EM to decouple the joint distribution of hazy and clear images, while the second stage leverages the pre-trained model to generate pseudo endpoints for semi-supervised training. We also introduce a Residual Difference Convolution (RDC) block to capture gradient-level details, improving texture and edge representation. Extensive experiments show that EM-B\textsuperscript{3}DM outperforms state-of-the-art methods in both qualitative and quantitative evaluations, demonstrating its superior performance with limited paired data. 
Future work could explore a transition towards unsupervised training strategies, with a continued emphasis on improving perceptual quality and result fidelity.


\clearpage
\bibliographystyle{ACM-Reference-Format}
\bibliography{samples/main}

\end{document}